\documentclass[10pt, a4paper,conference]{IEEEtran}
\usepackage{graphicx}
\usepackage{amsmath}
\usepackage{hyperref}
\graphicspath{ {./images/} }

\ifCLASSINFOpdf
\else
\fi
\usepackage{bm}
\begin{document}
%
\title{Deep transformation models: Tackling complex regression problems with neural network based transformation models}


\author{\IEEEauthorblockN{Beate Sick \IEEEauthorrefmark{1}}
\IEEEauthorblockA{EBPI, University of Zurich \& \\
IDP, Zurich University of Applied Sciences\\
Email: beate.sick@uzh.ch, sick@zhaw.ch}
\and
\IEEEauthorblockN{Torsten Hothorn}
\IEEEauthorblockA{EBPI, University of Zurich\\
Email: torsten.hothorn@uzh.ch}
\and
\IEEEauthorblockN{Oliver D{\"u}rr \IEEEauthorrefmark{1}} 
\IEEEauthorblockA{IOS, Konstanz University of Applied Sciences\\
Email: oliver.duerr@htwg-konstanz.de}
\and
\IEEEauthorblockA{\IEEEauthorrefmark{1} corresponding authors, contributed equally}
}


%


\maketitle

\begin{abstract}
We present a deep transformation model for probabilistic regression. 
Deep learning is known for outstandingly accurate predictions on complex data but in regression tasks it
is predominantly used to just predict a single number. This ignores the non-deterministic character of most tasks. Especially if crucial decisions are based on the predictions, like in medical applications, it is essential to
quantify the prediction uncertainty. The presented deep learning transformation model estimates the whole
conditional  probability  distribution,  which is the most thorough way to capture uncertainty about the outcome. 
We combine ideas from a statistical transformation
model  (most  likely  transformation)  with  recent  transformation
models from deep learning (normalizing flows) to predict complex
outcome distributions. The core of the method is a parameterized
transformation  function  which  can  be  trained  with  the  usual
maximum  likelihood  framework using gradient descent.  The  method  can  be  combined
with  existing  deep  learning  architectures.
For  small  machine  learning  benchmark  datasets,  we  report
state  of  the  art  performance  for  most  dataset  and  partly even
outperform  it.  Our  method  works  for
complex  input  data, which we demonstrate by  employing  a  CNN  architecture on image data.

\end{abstract}

%
\IEEEpeerreviewmaketitle

\section{Introduction}

Deep learning (DL) models are renowned for outperforming traditional methods in perceptual tasks like image or sound classification \cite{Lecun2015}. For classification tasks the DL model is usually treated as a probabilistic model, where the output nodes (after softmax) are interpreted as the probability for the different classes. The networks are trained by minimizing the negative log-likelihood (NLL). DL is also successfully used in regression tasks with complex data. However, regression DL models are often not used in a probabilistic setting and yield only a single point estimate for the expected value of the outcome. 

Many real world applications are not deterministic, in a sense that a whole range of outcome values is possible for a given input. In this setting, we need a model that does not only predict a point estimate for the outcome conditioned on a given input, but a whole conditional probability distribution (CPD). 

In many regression applications a Gaussian CPD is assumed, often with the additional
assumption that the variance of the Gaussian does not depend on the input
$x$ (homoscedasticity).  In the Gaussian homoscedastic cases only one node in
the last layer is needed and interpreted as the conditional mean $\hat{\mu}_{x}$
of the CPD $N(\hat{\mu}_{x}, \sigma^2)$.  In this case the NLL is
proportional to the mean squared error $\frac{1}{n} \sum_{i=1}^n(y_i -
\hat{\mu}_{x_i})^2$, which is one of the best known loss functions in
regression.  Minimizing the loss function yields the maximum likelihood
estimate of the parameter $\mu$ of the Gaussian CPD.  The parameter
$\sigma^2$ of the CPD is estimated from the resulting residuals $r_i=y_i -
\hat{\mu}_{x_i}$, an approach that only works in the homoscedastic case.  In
case of non-constant variance, a second output node is needed which controls
the variance.

The likelihood approach generalizes to other types of CPDs, especially if
the CPD is known to be a member of a parameterized distribution family with parameters conditioned on the input $x$.

\section{Related Work}
A simple parameterized distribution like the Gaussian is sometimes not flexible enough to model the
conditional distribution of the outcome in real-world applications.  A
straightforward approach to model more flexible CPDs is to use a mixture of
simple distributions.  Mixtures of Gaussian, where the mixing coefficients
and the parameters of the individual Gaussians are controlled by a neural
network, are known as neural density mixture networks and have been used for
a long time \cite{Bishop1994}.

Another way to achieve more complex distributions is to average over many
different CPDs, where each  CPD is a simple distribution.  This approach is
taken by  ensemble models \cite{NIPS2017_7219} or alternatively Bayesian
models.  A Bayesian deep learning treatment for regression has been done by
MC-Dropout \cite{Gal2015,Gal2017b}

Transformation models take a different approach to model a flexible
distributions.  Here, a parameterized bijective transformation function is
learned that transforms between the flexible target distribution of the
outcome $y$ and a simple base distribution of a variable $z$, such as a
Standard Gaussian.  The likelihood of the data can be easily determined from the base
distribution after transformation using the change of variable formula. 
This allows to determined the parameters of the transformation using the
maximum likelihood estimation.

\subsection{Normalizing flows}
On the “deep learning” side the normalizing flows (NF) have recently
successfully been used to model complex probability distributions of
high-dimensional data \cite{Rezende2015}; see \cite{Papamakarios2019} for an
extensive review.  A typical NF transforms from a simple base distribution
with density $f_z(z)$ (often a Gaussian) to a more complex target
distribution $f_y(y)$.  This is implemented by
means of a parameterized transformation function $h_{\theta}^{-1}(z)$.

A NF usually consists of a chain of simple transformations interleaved with
non-linear activation functions.  To achieve a bijective overall
transformation, it is sufficient that each transformation in the chain is 
bijective.  The bijectivity is usually enforced by choosing a strictly
monotone increasing transformation.  A simple but efficient and often used
transformation consists of a scale term $a(z)$ and a shift term $b(z)$:

\begin{equation}
    y=h_\theta^{-1}(z) = a(z) \cdot z - b(z).
\end{equation}

To enforce the required monotony of the transformation $a(z)$ needs to be positive.  Note that
the density in $y$ after transformation is determined by the change of
variable formula:

\begin{equation}
f_y(y)=f_z(h_\theta(y))|h_\theta'(y)|.
\end{equation}

The Jacobian-Determinant $|h_\theta'(y)|$ ensures that $f_y(y)$ is again normalized. The parameters in the transformation function, here  $(a(z), b(z))$ are controlled by the output of a neural network carefully designed so that the Jacobi-Determinant is easy to calculate \cite{Papamakarios2019}. To ensure that $a(z)$ is positive, the output of a network $\tilde a(z)$ is transformed
e.g.~by $a(z)=\exp(\tilde a(z))$. By chaining several such transformations with non-linear activation functions inbetween, a target distributions can be achieved which has not only another mean and spread, but also another shape than the base distribution.

NFs are very successfully used in modeling high dimensional unconditional distributions.  They can be used for data generation, like creating photorealistic samples human faces \cite{Kingma2018}. However, little research is done yet with respect to using NF in a regression-like setting, where the distribution of a low-dimensional variable $y$ is modeled conditioned on a possible high-dimensional variable $x$ \cite{trippe2018},\cite{Rothfuss}. In the very recent approach of \cite{Rothfuss} a neural network controls the parameters of a chain of normalizing flow, consisting of scale and shift terms combined with additional simple radial flows.

\subsection{Most Likely Transformation}
On the other side, transformation models have a long tradition in classical
statistics, dating back to the 1960’s starting with the seminal work of Box
and Cox on transformation models \cite{Box1964}.  Recently this approach has
been generalized to a very flexible class of  transformations called most
likely transformation (MLT) \cite{Hothorn2018}.  The MLT models a complex
distribution $p_y(y \mid x)$ of a one-dimensional random variable $y$ conditioned
on a possibly high-dimensional variable $x$.  While multivariate extensions
of the framework exists \cite{klein2019multivariate}, we restrict our treatment to the
practically most important case of a univariate response $y$.  As in normalizing
flows the fitting is done by learning a parameterized bijective
transformation $h_\vartheta(y \mid x)$ of $(y \mid x)$ to a simple distribution
$(z \mid x)=h_\vartheta(y \mid x) \sim \Phi$, such as a standard Gaussian
$\Phi=N(0,1)$.  The parameters are determined using the maximum likelihood
principle to the training data.  For continuous outcome
distributions, the change of variable theorem allows
computing the probability density $f_y(y \mid x)$ of the complicated target distribution
from the simple base distribution $f_z(z)$ via:

\begin{equation}
    f_y(y \mid x)=f_z(h(y \mid x)) \cdot |h'(y \mid x)|.
\end{equation}

As in the NF, the factor $|h'(y \mid x)|$ ensures the normalization of $f_y(y \mid
x)$. 
The key insight of the MLT method is to use a flexible class of polynomials,
e.g.~the Bernstein polynomials, to approximate the transformation function
\begin{equation}
\label{eq:mlt}
    h_\vartheta^{\tt MLT}(\tilde y \mid x) = \sum_{i=0}^M {\tt Be_i}(\tilde{y}) \frac{\vartheta_i(x)}{M+1}
\end{equation}
where $\tilde  y \in [0,1]$.  The core of this transformation is the
Bernstein basis of order $M$, generated by the $M+1$ Beta-densities  ${\tt
Be_i}(\tilde{y}) = f_{i+1, M-i+1}(\tilde y)$.  It is known that the
Bernstein polynomials are very flexible basis functions and uniformly
approximate every function in $y \in [0,1]$, see \cite{Hothorn2018} for a
further discussion.  The second reason for the choice of the Bernstein basis
is that in order for $h_\vartheta(\tilde y \mid x)$ to be bijective, a strict
monotone transformation of $\tilde y$ is required.  A strict increase of a
polynomial of the Bernstein basis can be guaranteed by simply enforcing that the
coefficients of the polynomial are increasing, i.e.  $\vartheta_0 <
\vartheta_1 < \ldots < \vartheta_M$. It should be noted that
nonparametric methods of the transformation function $h$ dominate
the literature, and smooth low-dimensional spline approaches only recently
received more attention; see \cite{Tian_Hothorn_Li_2020} for a comparison of
the two approaches.

A useful special case of the transformation function, is the so-called linear transformation function which has the following form:
\begin{equation}
\label{eq:LTM}
    h_\vartheta^{\tt LTM}(\tilde y \mid x) = h(\tilde y) -  \sum_{p=1}^P {\beta_p \cdot x_p} 
\end{equation}
This linear transformation model (LTM) is much less flexible than the transformation in Equation \ref{eq:mlt}. Its main advantage is the interpretability of the coefficients as in classical regression models. Linear transformation models have been used for a long time in survival analysis. For example a Cox model is a linear transformation model: If the base distribution is a minimum extreme value distribution, the coefficients $\beta_p$ are the log hazard ratios. If the base distribution is chosen to be the a standard logistic distribution then the coefficients $\beta_p$  are the log odds ratios, see \cite{Hothorn2018} for a detailed discussion.

\section{Method}
In the following, we combine the ideas of NF parameterized by neural networks and MLT. 
Our model consists of a chain of four transformations constructing the parametric transformation function:
\begin{equation}
\label{eq:MLT}
    z = h_\theta(y) = f_{3, \alpha, \beta } \circ f_{2, \vartheta_0, \ldots, \vartheta_M} \circ \sigma \circ f_{1, a, b}(y)
\end{equation}

\begin{figure}
\centering
\includegraphics[width=.95\linewidth]{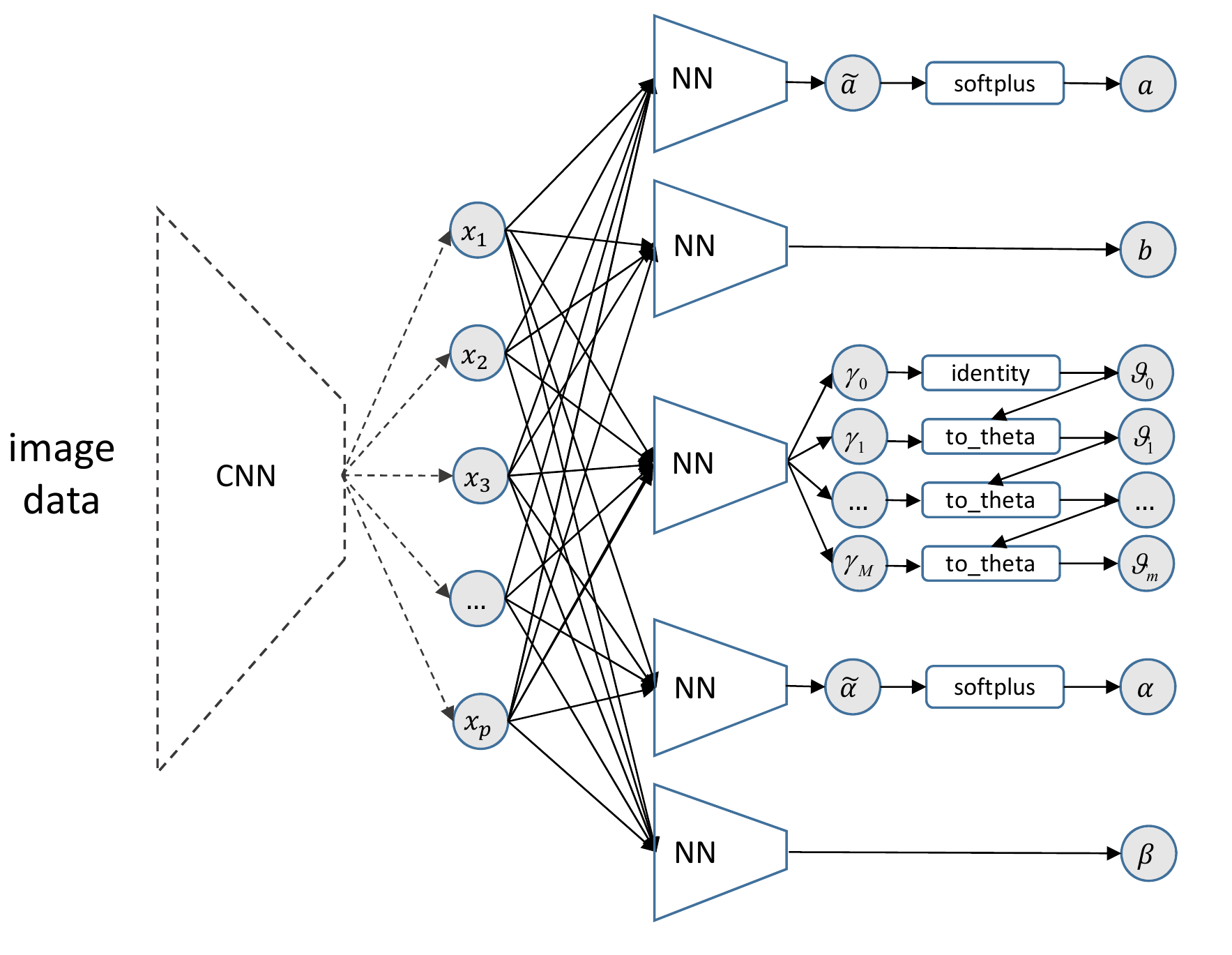}
\caption{The network architecture of our model which is trained end-to-end starting with the input x yielding the conditional parameters of the transformation functions. The dashed CNN part is only used in case of image input data. If some parameters should be not depended on 
$x$, the respective NN gets a constant number (one) as input. }
\label{fig:NN}
\end{figure}

Only the sigmoid transformation $\sigma$ has no parameters, whereas $ f_1,
f_2, f_3$ have learnable parameters.  In a regression task we allow in our
flexible model all parameters to change with $x$.  We use neural networks to
estimate the parameters (see Figure \ref{fig:NN}).  We describe the chain of
flows going from the untransformed variable $y$ to the standard normally
distributed variable $z$ (see Figure \ref{fig:fig_trafo}).  The MLT approach based on Bernstein polynomials
needs a target variable which is limited between $[0,1]$.  For our model, we
do not want to apriori define a fixed upper and lower bound of $y$ to
restrict its range.  Instead, we transform in a first flow $f_1$, $y$ to the
required range of $[0,1]$.  This is done by a scaled and shifted
transformation $f_1$ followed by a sigmoid function via (see Figure \ref{fig:fig_trafo}):
\begin{equation}
   \sigma \circ  f_1: \;\; \tilde{y} = \sigma(a(x) \cdot y - b(x)).
\end{equation}
The scale and shift parameters $(a, b)$ are allowed to depend on $x$ and are
controlled by the output of two neural networks with input $x$.  In order to
enforce $f_1$ to be monotonically increasing, we need to restrict  $a(x) >
0$.  This is enforced by a softplus activation of the last neuron (see
Figure \ref{fig:NN}).

The second transformation $f_2$ is the MLT transformation from
Equation~(\ref{eq:mlt}) consisting of a Bernstein polynomial with $M+1$
parameters $\vartheta_0 , \vartheta_1,\ldots \vartheta_M$ which can depend
on $x$ and are controlled by a NN with input $x$ and as many output nodes as
we have parameters. If the order $M$ of the Bernstein polynomial is $M=1$, then the $f_2$-transformation is linear and does not change the shape of the distribution. Therefore, for heteroscedastic linear regression a Bernstein polynomial of $M=1$ is sufficient when working with a Gaussian base distribution. But with higher order Bernstein polynomials we can achieve very flexible transformation that can e.g. turn a bimodal distribution into a Gaussian (see Figure \ref{fig:fig_trafo}).
To ensure a monotone increasing $f_2$, we enforce the
above discussed condition $\vartheta_0 < \vartheta_1 < \ldots < \vartheta_M$
by choosing $\vartheta_k = \vartheta_{k-1} + \exp(\gamma_k)$ for $k>0$ and
$\gamma_k$ being the $k$-th output of the network.  We set
$\vartheta_0=\gamma_0$ (see Figure~\ref{fig:NN}).

The final flow in the chain $f_3$ is again a scale and shift transformation
into the range of the standard normal (see Figure \ref{fig:fig_trafo}).  The parameters $(\alpha, \beta)$ are
also allowed to depend on $x$ and are controlled again by NNs, where the
softplus transformation is used to guarantee a positive $\alpha$ (see
Figure~\ref{fig:NN}).  First experiments showed that the third flow is not
necessary for the performance but accelerates training (data not shown).

The total transformation $ h_\theta(y): y \rightarrow z $ is given by
chaining the three transformations resulting in (see Figure trafo):

\begin{equation}
\label{eq:DLMLT}
\begin{split}
    z_x &= h_{\theta(x)}(y) = \alpha(x) \cdot h_{\vartheta(x)}^{\tt MLT}(\tilde y) - \beta(x) \\\
    &= \alpha(x) \left( \sum_{i=0}^M {\tt Be_i}(\sigma(a(x) \cdot y - b(x)) \frac{\vartheta_i(x)}{M+1} \right) - \beta(x).
\end{split}
\end{equation}

Our model has been implemented in R using Keras, TensorFlow, and TensorFlow Probability and can be found at: \url{https://github.com/tensorchiefs/dl_mlt}

\begin{figure}
\centering
\includegraphics[width=1.0\linewidth]{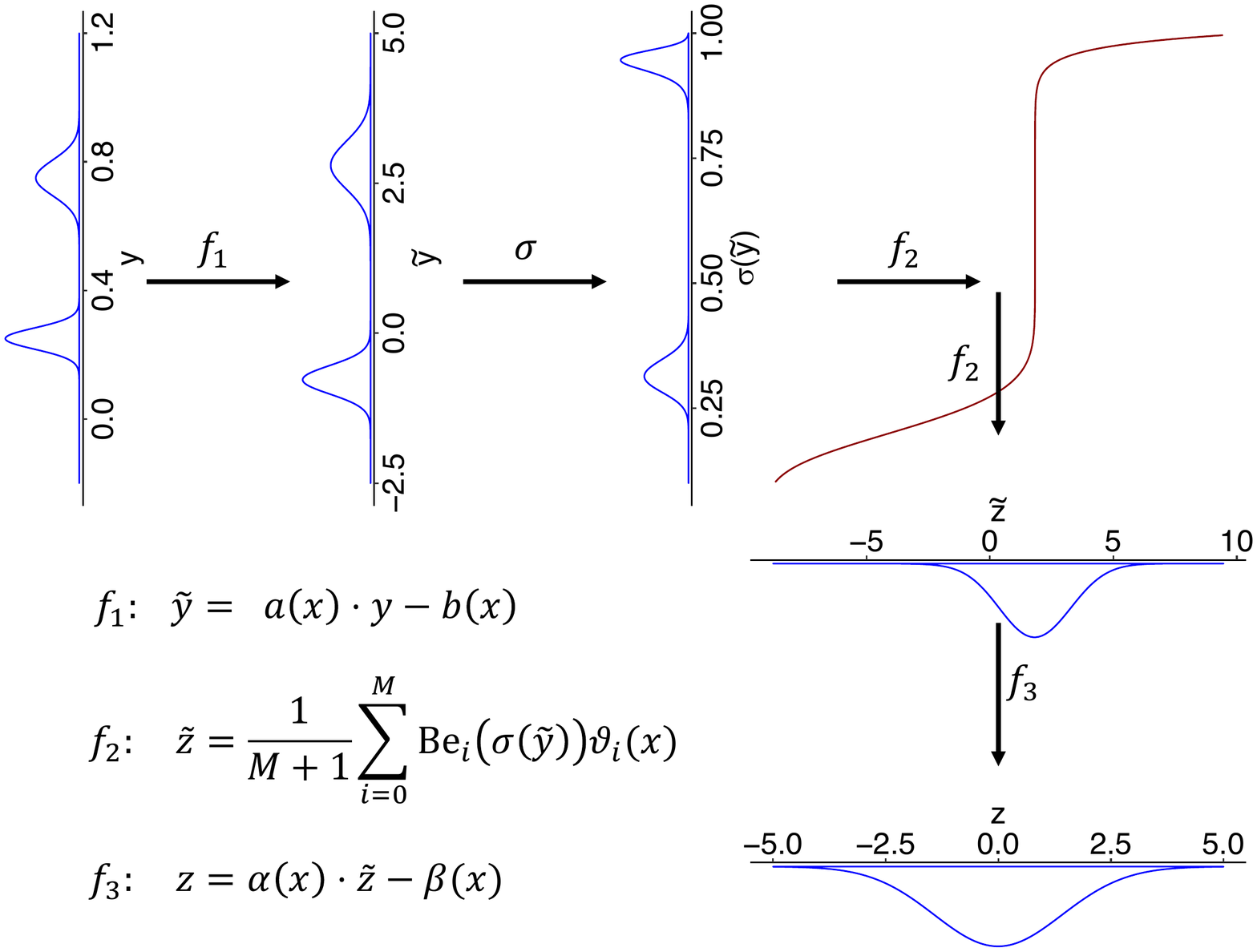}
\caption{Visual representation of the flow of Equation \ref{eq:MLT}. The flow transforms from a bimodal distribution, such as the CPD in Figure \ref{fig:fig_sin} at $x=2.5$, to a standard Gaussian. For $f_2$ where the change from a bimodal to unimodal distributions happens the transformation function is also on display.}
\label{fig:fig_trafo}
\end{figure}

\section{Experiments}
In order to demonstrate the capacity of our proposed approach, we performed three different experiments.

\subsection{One-Dimensional Toy Data}
\begin{figure}
\centering
\includegraphics[width=.7\linewidth]{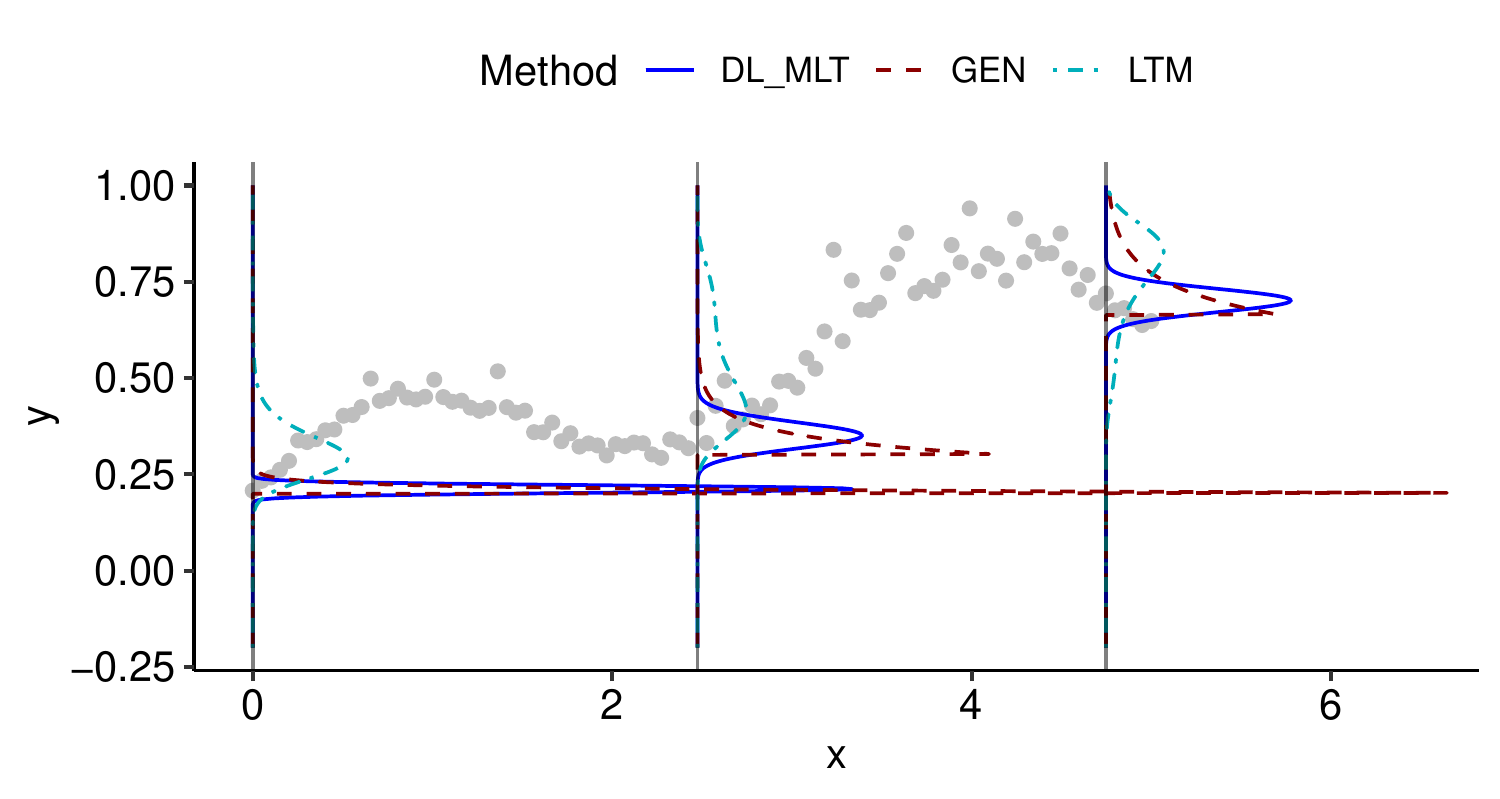}
\includegraphics[width=.7\linewidth]{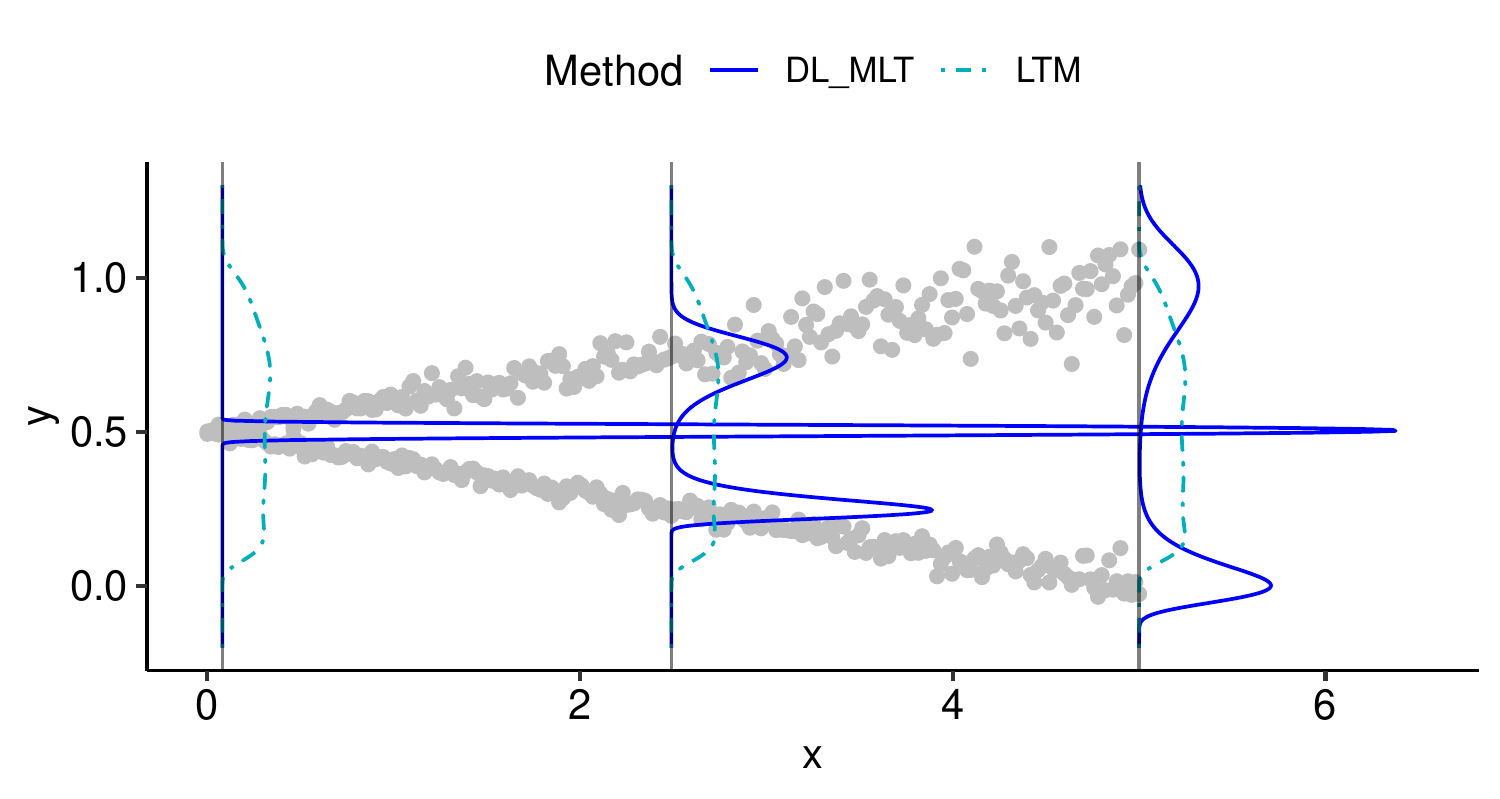}
\caption{Simple one-dimensional toy example with complex outcome
distributions.  Training data is shown as gray points and the estimated CPDs
for the linear transformation model (LTM) and the proposed DL\_MLT method are shown for two
data generating process as a dashed and solid lines, respectively.  In the
upper panel in addition the true CPD of the data generating process is shown
(dotted line).  The CPDs have been scaled down for plotting.
}
\label{fig:fig_sin}
\end{figure}

We use simulated toy data sets with a single feature and a single target
variable to illustrate some properties of the proposed transformation model. 
In the first example the data generating process is given by imposing
exponential noise with increasing amplitude on a linear increasing
sinusoidal (see Figure \ref{fig:fig_sin}).  For some $x$-values two fitted
CPDs are shown, one resulting from our NN based transformation model with
$M=10$ (DL\_MLT) Equation \ref{eq:DLMLT} and one from a less flexible linear transformation
model (LTM) Equation \ref{eq:LTM}. Figure \ref{fig:fig_sin} clearly demonstrates that the
complex transformation model  can, in contrast to the simple linear
transformation model, cope with an non-monotone $x$-dependency of the variance.  To
quantify the goodness of fit, we us the negative log-likelihood (NLL) of the
data (smaller is better), which is -2.00 for the complex and -0.85 for the
simple linear transformation model.

The second example, shown on the lower panel of Figure~\ref{fig:fig_sin}, is
a challenging bimodal distribution in which the spread between the modes
strongly depends on $x$ but the conditional mean of the outcome $y$ does not
depend on $x$.  Both models can capture the bimodal shape of the CPD, but
the LTM is not flexible enough to adapt its bimodal CPD with
changing $x$.  The proposed flexible transformation model DL\_MLT is able to also
take the $x$ dependency of the spread between the modes of the distribution
into account.

\subsection{UCI-Benchmark Data Sets}
\begin{table*}[!t]
\renewcommand{\arraystretch}{1.3}
\caption{Comparison of prediction performance (test NLL, smaller is better)  on regression benchmark UCI datasets. The best method for each dataset is bolded, as are those with standard errors that overlap with the standard errors of the best method.}
\label{table_example}

\centering
\begin{tabular}{r|c|ccccccc}
\hline
Data Set & N & DL\_MLT & NGBoost & MC Dropout & Deep Ensembles & Gaussian Process & MDN & NFN \\ \hline
Boston & 506 & \bf{2.42 $\pm$ 0.050} & \bf{2.43 $\pm$ 0.15} & \bf{2.46 $\pm$0.25} & \bf{2.41 $\pm$0.25} & \bf{2.37 $\pm$0.24} & \bf{2.49 $\pm$ 0.11} & \bf{2.48 $\pm$0.11} \\
Concrete & 1030 & 3.29 $\pm$ 0.02 & \bf{3.04 $\pm$ 0.17} & \bf{3.04 $\pm$0.09} & \bf{3.06 $\pm$0.18} & \bf{3.03 $\pm$0.11} & \bf{3.09 $\pm$ 0.08} & \bf{3.03 $\pm$0.13} \\
Energy & 768 & \bf{1.06 $\pm$ 0.09} & \bf{0.60 $\pm$ 0.45} & 1.99 $\pm$0.09 & 1.38 $\pm$0.22 & \bf{0.66 $\pm$0.17} & \bf{1.04 $\pm$ 0.09} & 1.21 $\pm$0.08 \\
Kin8nm & 8192 & -0.99 $\pm$ 0.01 & -0.49 $\pm$ 0.02 & -0.95 $\pm$0.03 & \bf{-1.20 $\pm$0.02} & -1.11 $\pm$0.03 & NA & NA \\
Naval & 11934 & \bf{-6.54 $\pm$ 0.03} & -5.34$\pm$ 0.04 & -3.80 $\pm$0.05 & -5.63 $\pm$0.05 & -4.98 $\pm$0.02 & NA & NA \\
Power & 9568 & \bf{2.85 $\pm$ 0.005} & \bf{2.79 $\pm$ 0.11} & \bf{2.80 $\pm$0.05} & \bf{2.79 $\pm$0.04} & \bf{2.81 $\pm$0.05} & NA & NA \\
Protein & 45730 & \bf{2.63 $\pm$ 0.006} & 2.81 $\pm$ 0.03 & 2.89 $\pm$0.01 & 2.83 $\pm$0.02 & 2.89 $\pm$0.02 & NA & NA \\
Wine & 1588 & \bf{0.67 $\pm$ 0.028} & 0.91 $\pm$ 0.06 & 0.93 $\pm$0.06 & 0.94 $\pm$0.12 & 0.95 $\pm$0.06 & NA & NA \\
Yacht & 308 & \bf{0.004 $\pm$ 0.046} & \bf{0.20 $\pm$ 0.26} & 1.55 $\pm$0.12 & 1.18 $\pm$0.21 & 0.10 $\pm$0.26 & NA & NA \\\hline
\end{tabular}

\end{table*}

To compare the predictive performance of our NN based transformation model with other state-of-the-art methods, we use nine well established benchmark data sets (see table bench) from the UCI Machine Learning Repository. Shown is our approach (Deep\_MLT), NGBoost \cite{Duan2020}, MC Dropout \cite{Gal2015}, Deep Ensembles \cite{NIPS2017_7219}, Gaussian Process \cite{Duan2020}, noise regularized Gaussian Mixture Density Networks (MDN) and normalizing flows networks (NFN) \cite{Rothfuss}. As hyperparameters for our Deep\_MLT model, we used $M=10$ for all datasets and a
$L_2$-regularization constants of 0.01 for the smaller datasets  (Yacht, Energy, Concrete, and Boston) no regularization was used in the other datasets.
To quantify the predictive performance, we use the negative log-likelihood (NLL) on test data. The NLL is a strictly proper score that takes its minimum if and only if the predicted CPD matches the true CPD and is therefore often used when comparing the performance of probabilistic models.

In order to compare with other state-of-art models, we follow the protocol from Hernandez and Lobato \cite{Hernandez-Lobato2015} that was also employed by Gal \cite{Gal2015}. For validating our method, the benchmark data sets were split into several folds containing training data (90 percent) and test-data (10 percent).  We downloaded the specific folds as used in  \cite{Gal2015} from \url{https://github.com/yaringal/DropoutUncertaintyExps}. We determined the hyperparameters of our model using 80 percent of the training data, keeping 20 percent for validation. The only preprocessing step has been a transformation to $[0,1]$  of $x$ and $y$ based on the training data. In contrast to other approaches in
Table~(\ref{table_example}) like \cite{Gal2015} and \cite{Duan2020}, we choose only one set of hyperparameters for each dataset and do not adapt the hyperparameters for each individual run separately. We can do this since our model has only few and quite stable hyperparameters. Specifically, we verified that the initial choice of the number of layers in the network is appropriate (no tuning has been done). We further choose the parameter of the
$L_2$-regularization of the networks and the order $M$ of the Bernstein polynomials. After choosing the hyperparameters, we trained the same model again on the complete training fold and determined the NLL on the corresponding test fold. This procedure allowed us to compare the mean, the standard error of the test NLL with reported values from the literature.

\begin{figure}[ht!]
\centering
\includegraphics[width=1.0\linewidth]{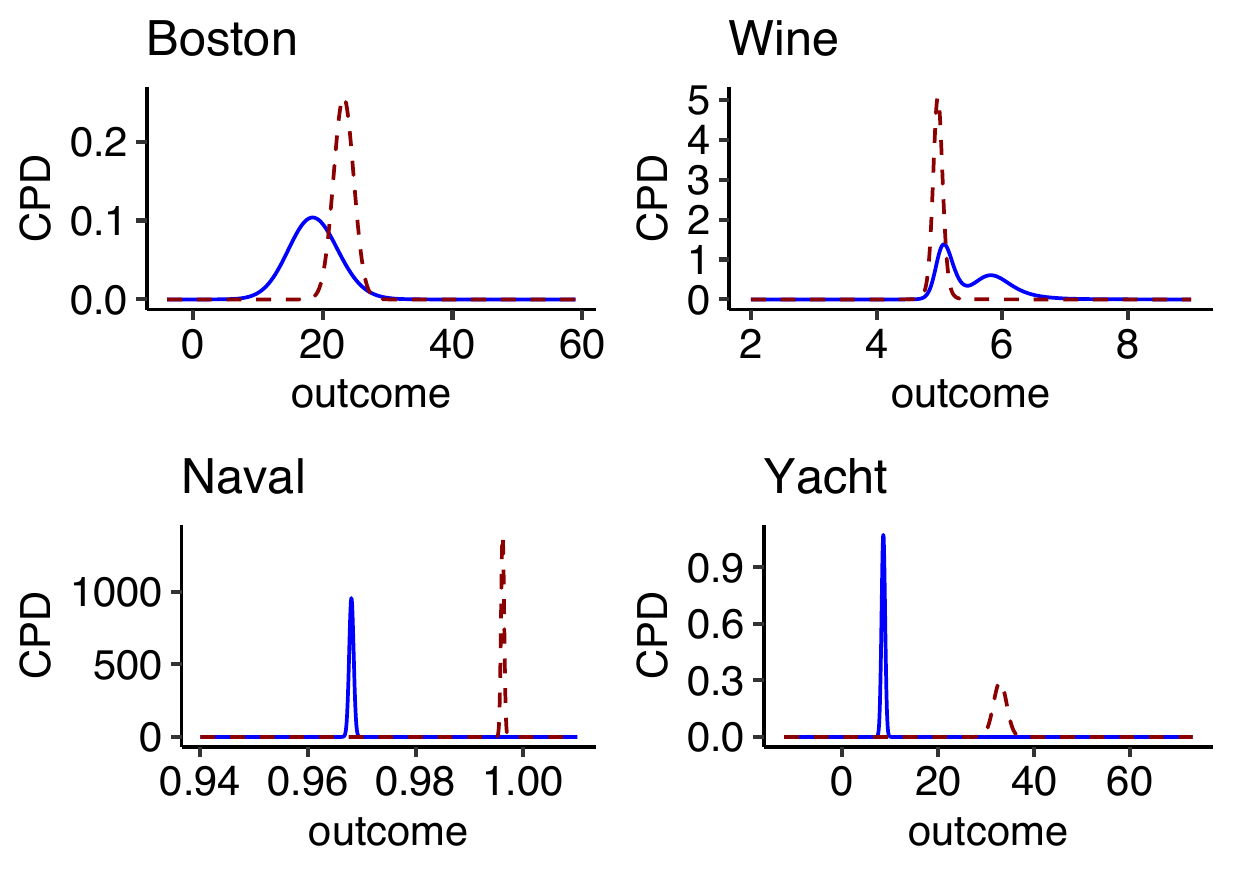}
\caption{Predicted CPDs for test data in the benchmark data Boston, Wine, Naval, and Yacht are shown. For each data set two test observations with commonly observed CPD shapes were picked (depicted with solid and dashed lines respectively). }
\label{fig:cpdwine}
\end{figure}

The results in Table~(\ref{table_example}) show that our DL\_MLT model yields, overall, a competitive performance with other state-of-the-art models. In the two tasks (Naval and Wine) the transformation model clearly outperforms all existing models. We followed up on the reasons and visually inspected the predicted CPDs in the different tasks. For Naval, we found that the model needs extremely long training periods, approximately 120’000 iterations compared to approximately 20’000, for the other data sets. When trained for such as a period it’s CPD is a very narrow single spike (see lower left panel in Figure \ref{fig:cpdwine}). We speculate that the reason for worse NLL of the other models is that they have not been trained long enough.  For the other data-sets, were we outperformed the existing methods, the striking finding was, that for many instances very non-Gaussian CPDs are predicted. For the Wine dataset the predicted CPDs are not always uni-model but quite often show a multi-modal shape (see upper right panel in Figure \ref{fig:cpdwine}). This makes perfectl sense, since the target variable in the dataset, the subjective quality of the wine is an integer value ranging from 0 to 10\footnote{Here we treat the values as continuous quantities as done in the reference methods, when taking the correct data type into account.}. Hence, if the model cannot decide between two levels, it predicts a bimodal CPD , which shows that it can learn to a certain extend the discrete structure of the data (see upper right panel in Figure \ref{fig:cpdwine}). 

To investigate if we need more flexibility to model this kind of multi-modal outcome distributions, we refitted the Wine data with $M=20$ and obtained an significantly improved test NLL of $0.40 \pm 0.025$. For the other data-sets the underlying cause of the better performance of our DL\_MLT model is unclear to us. 

\subsection{Age Estimation from Faces}
In this experiment, we want to demonstrate the usefulness of the proposed NN based transformation model for a typical deep learning context dealing with complex input data, like images. For this purpose, we choose the UTKFace dataset containing $N=23’708$ images of cropped faces of humans with known age ranging from 1 to 116 years \cite{Zhang}. The spread of the CPD indicates the uncertainty of the model about the age. We did not do any hyperparameter search for this proof of concept and used 80 percent of the data for training. The color images of size $(200, 200, 3)$ were downscaled to $(64,64, 3)$ and feed into a small convolutional neural network (CNN) comprising three convolutional layers with 141840 parameters, as indicated in Figure \ref{fig:NN}.

\begin{figure}[ht!]
\centering
\includegraphics[width=\linewidth]{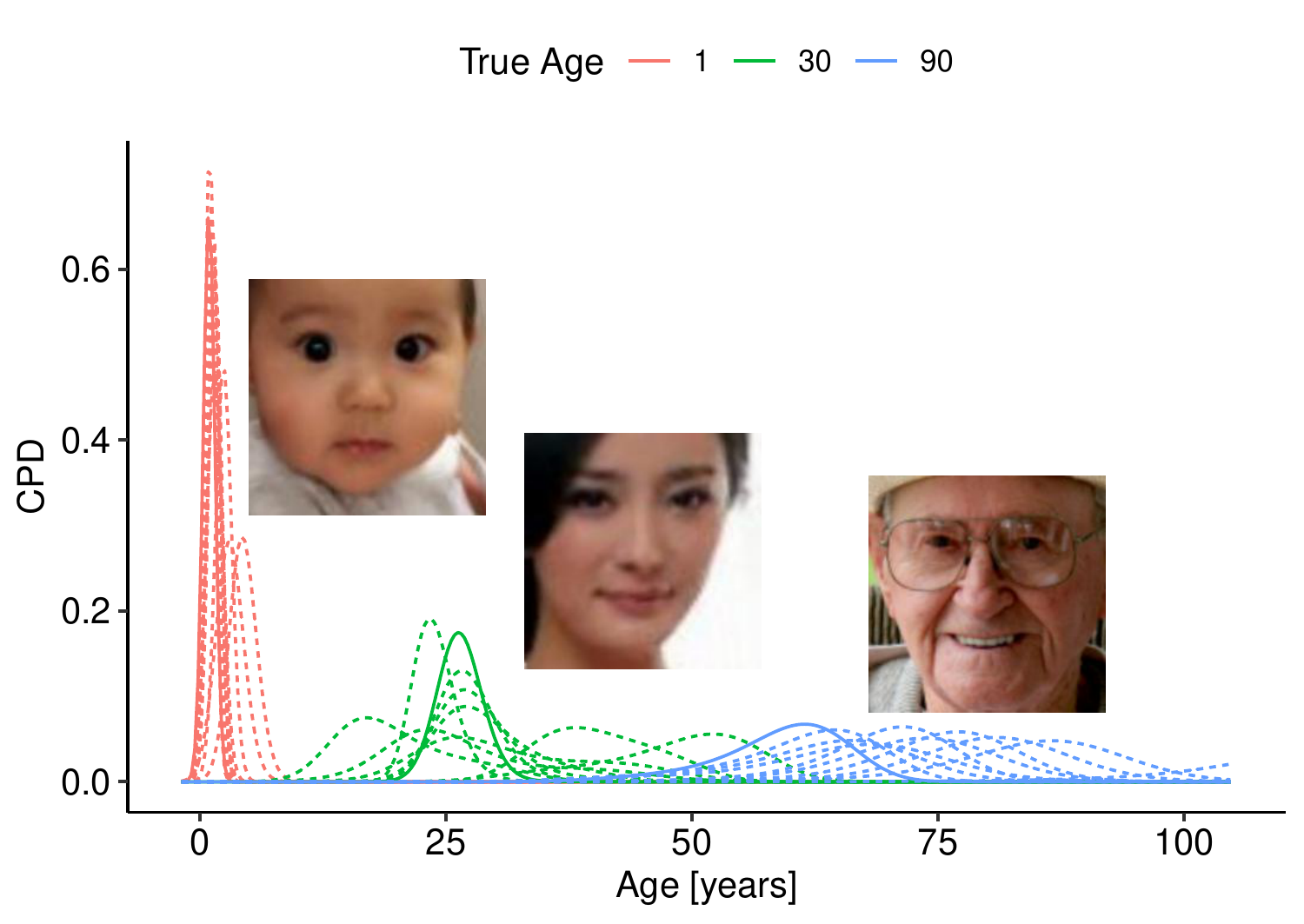}
\caption{CPD for different ages. Shown are 10 randomly picked results of the test set for different true ages (1, 30, and 90 years), the shown images correspond to the solid lines.}
\label{fig:age}
\end{figure}

Estimating the age is a challenging problem. First, age is non negative. Second, depending on the age, it is more or less hard to estimate the age. As humans we would probably be not more than two years off when estimating the age of a one year old infant, but for a thirty or ninety year old person the task is much harder.  Therefore, we expect that also a probabilistic model will predict CPDs with different spreads and shapes. Usually this problem is circumvented by transforming the age by
e.g.~a log transformation. Here, we did not apply any transformation to the raw data and let the network find to correct CPD in an end-to-end fashion. After training for 15’000 iterations, a NLL of 3.83 on the test-set has been reached. Since we did not find a suitable probabilistic benchmark for the dataset, we show in
Figure~\ref{fig:age} some typical results for the CPD for people of different ages.

The results of the model meet our expectations yielding narrow CPDs for infants, while for older people the spread of the CPD broadens with increasing age.

\section{Conclusion and outlook}
We have demonstrated that deep transformation models are a powerful approach
for probabilistic regression tasks, where for each input $x$ a conditional
probability distribution on $y$ is predicted.  Such probabilistic models are
needed in many real-world situations where the available information does
not deterministically define the outcome and where an estimate of the
uncertainty is needed.

By joining ideas from statistical transformation models and deep normalizing
flows the proposed model is able to outperform existing models from both
fields when the shape of the conditional distribution is complex and far
away from a Gaussian.  

Compared to statistical transformation models the proposed deep
transformation model does not require predefined features.  It can be
trained in an end-to-end fashion from complex data like images by prepending
a CNN.

Though mixture density networks \cite{Bishop1994} are able to
handle non-Gaussian CPDs they often tend to overfit the data and require a
careful tuning of regularization to reach a good performance
\cite{Rothfuss}.  Since deep transformation models estimate a smooth,
monotonically increasing transformation function, they always yield smooth
conditional distributions.  Therefore, we need only mild regularization in
cases where the NN for the parameter estimation would overfit the train
data.  For the UCI-datasets, we empirically observed the need for mild
regularization only in cases the size of the dataset is below 1500 points.

The proposed model can, in principle, be extended to higher dimensional
cases of the response variable $y$, see
\cite{klein2019multivariate} for a statistical treatment.  The fact that our
model is very flexible and does not impose any restriction on the modeled
CPD, allows it to learn the restrictions of the dataset, like age being
positive, from the data.  For limited training data that can lead to
situations where some probability is also assigned to impossible outcomes,
such as negative ages.  It is possible to adapt our model to respect these kind of limitations or to predict e.g. discrete CPDs. We will tackle this in future research. For discrete data we will use the appropriate likelihood that is described in terms of the corresponding discrete density $\Phi(h(y \mid x)) - \Phi(h(y-\mid x))$, where $y$ is the observed value and $y-$ the largest value of the discrete support with $y- < y$.  For the Wine dataset for example, where the outcome is an ordered set of ten answer categories, such an approach would result in a discrete predictive model matching the outcome $y$. At the moment, we also did not incorporate the
ability to handle truncated or censored data which is possible with
statistical transformation models, see \cite{Hothorn2018}. For the Boston dataset for example, all values with $y = 50$ are in fact right-censored, thus the probability $1 - \Phi(h(50 \mid x))$ would be the correct contribution to the likelihood.

Another limitation of the proposed flexible transformation NN is its black box
character.  Picking special base distributions, such as the standard
logistic or minimum extreme value distributions, and using linear transformation models should
allow to disentangle and describe the effect of different input features by
means of mean shifts, odds ratios, or hazard ratios.  This can be translated
into interpretable deep learning transformation models, which we plan to do in the near
future.

\section*{Acknowledgment}
The authors would like to thank Lucas Kook, Lisa Herzog, Matthias Hermann, and Elvis Murina for fruitful discussions.

\bibliographystyle{IEEEtran}
\bibliography{main.bib}
%



\end{document}